\documentclass[letterpaper, 10 pt, conference]{ieeeconf}
\IEEEoverridecommandlockouts
\usepackage{graphicx} 
\usepackage[export]{adjustbox}

\usepackage{amsmath, amsfonts, amssymb}
\usepackage{url}
\usepackage{algorithm}
\usepackage{algpseudocode}
\usepackage{subcaption}

\usepackage{mathtools} 

\renewcommand{\Re}{\mathbb{R}}
\newcommand{\Expectation}{\mathop{{}\mathbb{E}}}
\newcommand{\argmin}{\mathop{{}\mathrm{argmin}}}

\usepackage{amsthm}  

\newtheorem{lemma}{Lemma}
\newtheorem{theorem}{Theorem}
\newtheorem{proposition}{Proposition}
\newtheorem{remark}{Remark}

\allowdisplaybreaks[4]

\usepackage{flushend}

\makeatletter
\let\NAT@parse\undefined
\makeatother
\usepackage{hyperref}
\hypersetup{
   colorlinks=true,
    linkcolor= blue,
    allcolors=blue,
    citecolor = blue,
    filecolor=black,
    urlcolor=blue,
    }

\title{Active Learning with Dual Model Predictive Path-Integral Control for Interaction-Aware Autonomous Highway On-ramp Merging}

\author{Jacob Knaup$^{1}$,
Jovin~D'sa$^{2}$,
Behdad~Chalaki$^{2}$,
Tyler Naes$^{2}$,\\
Hossein~Nourkhiz Mahjoub$^{2}$,
Ehsan~Moradi-Pari$^{2}$,
Panagiotis Tsiotras$^{1}$
\thanks{\hangindent=0.5cm \scriptsize$^1$Institute for Robotics and Intelligent Machines, Georgia Institute of Technology,  Atlanta, GA 30332 USA. (email: jacobk@gatech.edu)
This work was conducted during J. Knaup's internship at Honda Research Institute,USA. }
\thanks{\hangindent=0.5cm \scriptsize$^2$Honda Research Institute USA, Inc. (email: jovin{\_}dsa@honda-ri.com, behdad{\_}chalaki@honda-ri.com)}
}

\begin{document}

\maketitle

\begin{abstract} 
  
   Merging into dense highway traffic for an autonomous vehicle is a complex decision-making task, wherein the vehicle must identify a potential gap and coordinate with surrounding human drivers, each of whom may exhibit diverse driving behaviors. 
   Many existing methods consider other drivers to be dynamic obstacles and, as a result, are incapable of capturing the full intent of the human drivers via this passive planning. 
   In this paper, we propose a novel dual control framework based on Model Predictive Path-Integral control to generate interactive trajectories. 
   This framework incorporates a Bayesian inference approach to \emph{actively} learn the agents' parameters, i.e., other drivers' model parameters. 
   The proposed framework employs a sampling-based approach that is suitable for real-time implementation through the utilization of GPUs. 
   We illustrate the effectiveness of our proposed methodology through comprehensive numerical simulations conducted in both high and low-fidelity simulation scenarios focusing on autonomous on-ramp merging.

  
\end{abstract}

\section{Introduction}

Autonomous driving, similar to many problems arising in robotics, requires optimal and safe decision-making in the presence of uncertainty about other agents' and humans' behavior \cite{claussmann2019review}. Since every person may have their own unique driving style, an autonomous vehicle must learn a unique model for each other driver and then plan its trajectory accordingly in order to safely and efficiently interact with them \cite{song2021surrounding}.
The task of highway on-ramp merging has been shown to be an especially challenging task for autonomous vehicles and even human drivers \cite{rios2016survey}, \cite{zgonnikov2020should}. 
In congested traffic scenarios, the merging vehicle must successfully negotiate with other drivers to find someone who will yield to them and open a gap to merge into, while also dealing with constraints such as the end of the merge-lane \cite{fernandez2021highway}.

There are different ways to approach these scenarios. 
For example, \cite{wang2021deep} uses an end-to-end learning-based approach in which deep reinforcement learning (RL) is used to learn an optimal policy for the merging vehicle.
Meanwhile, \cite{schuurmans2023safe} utilizes a stochastic model predictive control (SMPC) framework, wherein the authors formulate a scenario tree of possible traffic vehicle actions and solve the resulting nonlinear program (NLP). The authors of \cite{liu2022gaussian, evens2022learning} take a similar approach, but use Gaussian Process regression and a stochastic game theory approach, respectively, to model the unknown behavior of the other vehicles.
Finally, references \cite{kimura2022decision, albarella2023hybrid} take a hierarchical approach in which a high-level RL algorithm is trained to provide a reward or policy to a low-level MPC algorithm used for trajectory generation.

Selecting a fully learning-based framework for autonomous agents' decision-making can raise concerns due to its lack of transparency and interpretability \cite{lubars2021combining}, especially in safety-critical scenarios. 
Thus, in this work, we propose an interpretable model-based stochastic optimal control framework.
%
However, learning an \emph{interaction-aware} model for the behavior of a given driver in highly interactive scenarios is a challenging problem on its own. 
One way to achieve this is to probe the driver by specific actions of the autonomous car and observe how they respond \cite{hu2022active}. 
However, this involves a trade-off between exploring many potential yielding drivers and attempting to exploit the most promising drivers. 
Therefore, a \emph{dual control paradigm} is extremely attractive in these scenarios \cite{nair2022stochastic}. 
In a dual control framework, the stochastic optimal control policy is designed with regards to not only the current uncertainty in the system model, but also it takes into account how the future actions of the autonomous vehicle and the surrounding drivers will provide new information to better inform its model. 
Thus, a dual control framework \emph{actively learns} new information about other drivers while accomplishing the control objectives. 

Due to the real-time execution requirements, since the problem of highway merging involves dynamic obstacles, and since an interaction-aware driver model is inherently nonlinear, the gradient-free Model Predictive Path-Integral control (MPPI) algorithm is an attractive choice for this problem \cite{williams2017information}.
MPPI relies on randomly sampling control sequences and evaluating the cost of the corresponding predicted trajectories to compute the optimal control sequence as a weighted average of the sampled actions.

Several stochastic variations of MPPI have been previously proposed in the literature. For example, in \cite{abraham2020model}, the authors sample model parameter realizations in addition to control sequences to design a control sequence that is robust to potential model variations. Uncertainty-averse MPPI pairs MPPI with a mixture density network (a neural network trained to output the parameters of a Gaussian mixture model) and adds a term to the cost function corresponding to the degree of uncertainty corresponding to a given trajectory in order to penalize control sequences with a high degree of uncertainty \cite{arruda2017uncertainty}. Finally, Risk-aware MPPI (RA-MPPI) samples disturbance realizations in addition to control actions and then approximately computes the Conditional Value-at-Risk (CVaR) for each control sequence using the corresponding sampled trajectories and penalizes the control sequences that result in a high CVaR \cite{yin2023risk}. However, although there are several stochastic formulations of MPPI, there are no existing dual control formulations.

The contributions of this paper are twofold. First, we propose Dual Model Predictive Path-Integral Control (DMPPI), which extends MPPI by sampling model parameters and computing the expected cost over the \emph{expected future} distribution of the parameters as opposed to the current parameter distribution. 
Second, we demonstrate the performance of our approach on a realistic highway on-ramp merging scenario featuring congested traffic using the high-fidelity vehicle simulation software IPG-CarMaker \cite{carmaker2021reference}, thus, demonstrating the real-time computational capabilities of the algorithm in a multi-agent application. 
To the best of our knowledge, this is the first dual control formulation of MPPI and the first application of MPPI to autonomous highway merging.

\section{Problem Formulation}

We consider a general nonlinear stochastic system as
\begin{align} \label{eq:sys}
    x_{t+1} &= f(x_{t}, u_{t}, \bar{\theta}) + w_t,
\end{align}
where $x_t \in \Re^{n_x}$ and $u_t \in \Re^{n_u}$ denote the state and control action at time step $t\in\mathbb{N}$, respectively, while $\bar{\theta} \in \Theta \subseteq \Re^{n_p}$ is a vector of constant but unknown parameters. Let $w_t \in \Re^{n_x}$ be an i.i.d. random disturbance with corresponding density $a(w_t) = \mathcal{N}(0, \Sigma_w)$.
System \eqref{eq:sys} can also be expressed using the lifted dynamics given by
    $x_k = g_{k-t}(x_t, u_{t:k-1}, \bar{\theta}, w_{t:k-1})$,
for $k > t$ where $u_{t:k-1} = \{u_t, u_{t+1}, \dots, u_{k-1}\}$ and $w_{t:k-1} = \{w_t, w_{t+1}, \dots, w_{k-1}\}$. Given that the future state is contingent upon the realizations of the stochastic variable $w_t$, the dynamics can be expressed in terms of the conditional probability density, given by
\begin{align} \label{eq:sys_pdf}
    x_k \sim h_{k-t}(x_k | x_t, u_{t:k-1}, \bar{\theta}).
\end{align}

Associated with \eqref{eq:sys}, the cost function for the control horizon of length $N\in \mathbb{Z}^+$ is given by
\begin{align} \label{eq:cost_function}
    J(x_{t+1:t+N}) &= \phi(x_{t+N}) + \sum_{k=t+1}^{t+N-1} \ell_k(x_k),
\end{align}
where $x_{t+1:t+N} = \{x_{t+1}, x_{t+2}, \dots, x_{t+N}\}$, $\phi(\cdot)$ is the terminal cost, and $\ell_k(\cdot)$ the stage cost at time step $k$.
Our goal is to derive an optimal control sequence $u_{t:t+N-1|t} = \{u_{t|t}, u_{t+1|t}, \dots, u_{t+N-1|t}\}$, where $u_{k|t}$ is the action planned for time step $k$ at time step $t$, by solving the following optimization problem 
 \begin{align} \label{prob:simplified_ocp}
    \min_{u_{t:t+N-1|t}}~S(x_t, u_{t:t+N-1|t}, \bar{\theta}),
 \end{align}
where $S(\cdot)$ is defined as
    $S(x_t, u_{t:t+N-1|t}, \bar{\theta}) = \Expectation_{w_{t:t+N-1} \sim a}[J(g_1(x_t, u_{t|t}, \bar{\theta}, w_t),\allowbreak \dots,\allowbreak
    g_N(x_t, u_{t:t+N-1|t},\allowbreak \bar{\theta},\allowbreak w_{t:t+N-1}))]$.
Nonetheless, due to the unknown nature of $\bar{\theta}$, a direct solution to \eqref{prob:simplified_ocp} is not feasible. Instead, we model $\bar{\theta}$ by $\theta \sim b(\theta)$, with $b(\cdot)$ denoting the probability distribution that represents our belief of $\bar{\theta}$.
Rather than relying on a fixed belief distribution, we leverage Bayesian inference to estimate $\bar{\theta}$ online. 
Hence, the prior distribution is denoted as $b_0 = b(\theta)$, with the omission of the argument $\theta$ for brevity. The belief distribution is then updated based on the observed state as follows
\begin{align} \label{eq:belief_dynamics}
    b_{t+1}(\theta) &= b(\theta | x_{0:t+1}, u_{0:t})
    \propto b_{t}(\theta) h_{1}(x_{t+1} | x_{t}, u_{t}, \theta), 
\end{align}
for $ t = 0, 1, \dots$, where \eqref{eq:belief_dynamics} is derived from Bayes' theorem (e.g., as in \cite{speekenbrink2016tutorial}).

The key question now concerns the approximation of \eqref{prob:simplified_ocp} using \eqref{eq:belief_dynamics}. One straightforward method, denoted as Certainty Equivalence MPPI (CE-MPPI), is to, instead of \eqref{prob:simplified_ocp}, solve
\begin{align} \label{prob:ce_mppi_ocp}
    \min_{u_{t:t+N-1|t}}~&S(x_t, u_{t:t+N-1|t}, \Expectation_{\theta \sim b_t}[\theta]),
\end{align}
where we optimize with respect to the current expectation of the parameter using the previous data up to time step $t$. Another approach, similar to that in \cite{abraham2020model}, we refer to as Ensemble MPPI (EMPPI), which solves 
\begin{align}  \label{prob:emppi_ocp}
    \min_{u_{t:t+N-1|t}}~&\Expectation_{\theta \sim b_t}[S(x_t, u_{t:t+N-1|t}, \theta)],
\end{align}
through minimizing the expected cost over the current belief distribution of parameters at time $t$.
However, both of the aforementioned methods fail to take into account the potential acquisition of future information within the planning horizon. In the event that additional information is acquired during the execution of the trajectory, resulting in the convergence of $b(\cdot)$ towards the true value $\bar{\theta}$, incorporating this new knowledge will result in a smaller deviation between the computed plan and the optimal solution of \eqref{prob:simplified_ocp}.
\section{Proposed Approach}

\subsection{Dual Control Formulation}
The proposed approach takes into consideration the information gain in the planning horizon by incorporating a Bayesian update step, thus establishing an implicit dual control framework. 
While the optimization problem \eqref{prob:emppi_ocp} aims to minimize the expected cost based on the current belief state, our objective is to minimize the expected cost considering future belief states.

Therefore, we formulate the stochastic optimal control problem with the dual control framework as follows 
\begin{align} \label{prob:ideal_dc}
    &\min_{u_{t:t+N-1|t}} \bar{S}(x_t, u_{t:t+N-1|t}), \nonumber\\
    &\hspace{-3mm} \text{where }\bar{S}(x_t, u_{t:t+N-1|t}) \nonumber\\
    &= \Expectation_{\substack{\theta \sim b_{t+N},\\ w_{t:t+N-1} \sim a}}[\phi(g_N(x_t, u_{t:t+N-1|t}, \theta, w_{t:t+N-1}))] \nonumber\\
    &+ \sum_{k=t+1}^{t+N-1} \Expectation_{\substack{\theta \sim b_{k},\\ w_{t:k-1} \sim a}}[\ell(g_{k-t}(x_t, u_{t:k-1|t}, \theta, w_{t:k-1}))],
\end{align}
which computes the expected cost at each time step over the belief distribution using all information available at that time-step.

However, since the future belief distributions $b_k$ for $k > t$ cannot be calculated ahead of time as the future state realizations are unknown, the control policy resulting from Problem \eqref{prob:ideal_dc} is not causal. We instead use the approximate predicted future belief distributions $\hat{b}_{k|t}$, given by
\begin{align} \label{eq:pred_belief}
    &\hspace{-3mm}\hat{b}_{k+1|t}(\theta) \\
    &\hspace{-5mm}\propto \hat{b}_{k|t}(\theta) h_{k-t}
    \big(\Expectation_{\substack{\tilde{\theta} \sim b_{t},\\ w_{t:k} \sim a}}\left[g_{k+1}(x_t, u_{t:k}, \tilde{\theta}, w_{t:k})\right] | x_t, u_{t:k}, \theta\big), \nonumber
\end{align}
%
for all $k = t, \dots, t+N-1$ and where $\hat{b}_{t|t} = b_t$.

Thus, \eqref{eq:pred_belief} yields the \emph{causal} optimal control problem as
\begin{align} \label{prob:proposed}
    &\min_{u_{t:t+N-1|t}} \hat{S}(x_t, u_{t:t+N-1|t}), \nonumber\\
    &\hspace{-3mm} \text{where } \hat{S}(x_t, u_{t:t+N-1|t}) \nonumber\\
    &= \Expectation_{\substack{\theta \sim \hat{b}_{t+N|t},\\ w_{t:t+N-1} \sim a}}[\phi(g_N(x_t, u_{t:t+N-1|t}, \theta, w_{t:t+N-1}))] \nonumber\\
    &+ \sum_{k=t+1}^{t+N-1} \Expectation_{\substack{\theta \sim \hat{b}_{k|t},\\ w_{t:k-1} \sim a}}[\ell(g_{k-t}(x_t, u_{t:k-1|t}, \theta, w_{t:k-1}))],
\end{align}
which computes the expected cost at each time step over the predicted future belief distributions.

\begin{lemma}
    The solution to Problem \eqref{prob:proposed} is a causal control policy. That is, the optimal control applied at time $t$ only depends on information available at or before time $t$.
\end{lemma}
\begin{proof}
    The proof follows trivially from 
    \eqref{prob:proposed} which computes the expected cost using the current state $x_t$ and the planned future control sequence $u_{t:t+N-1|t}$ over the predicted future belief distributions. As seen in \eqref{eq:pred_belief}, the predicted belief dynamics are conditioned only on the current state $x_t$ and planned control sequence $u_{t:t+N-1|t}$. This is in contrast to \eqref{prob:ideal_dc} which employs the true belief dynamics given by \eqref{eq:belief_dynamics} that depend, however, on the future state realizations.
\end{proof}

\subsection{Information-Theoretic MPPI}

In practical applications, we address the computation of expectations in \eqref{eq:pred_belief} and \eqref{prob:proposed} through sampling-based approximations, as discussed further in Section \ref{sec:sampling}. 
The remaining challenge lies in solving Problem \eqref{prob:proposed}.
Given that the optimization problem over the control actions is, in general, non-convex, we utilize MPPI \cite{williams2018information}, a sampling-based optimal control framework to solve \eqref{prob:proposed}.

Following the information-theoretic derivation of MPPI \cite{williams2018information}, we first define a distribution from which we can sample candidate control sequences $u_t \sim \mathcal{N}(\bar{u}_t, \Sigma_u),$
where $\mathcal{N}(\bar{u}_t, \Sigma_u)$ is a multivariate normal distribution with mean $\bar{u}_t \in \Re^{n_u}$ and covariance $\Sigma_u \succ \mathbf{0}_{n_u \times n_u}$. Then, we may define the distribution $Q_{\bar{u}_{t:t+N-1|t}, \Sigma_u}$ for the control sequence $u_{t:t+N-1|t}$ with corresponding density given by
    $u_{t:t+N-1|t} \sim q(u_{t:t+N-1|t} | \bar{u}_{t:t+N-1|t}, \Sigma_u)$,
where $\bar{u}_{t:t+N-1|t} = \{\bar{u}_{t|t}, \bar{u}_{t+1|t}, \dots, \bar{u}_{t+N-1|t}\}$ is the mean and $\Sigma_u$ is the covariance. 

\begin{proposition}[\cite{williams2018information}]
    Control actions sampled from the optimal distribution $Q^{\ast}$, with corresponding density
    \begin{align*}
        q^{\ast}(u_{t:t+N-1|t}) = \frac{1}{\eta}\exp(-\dfrac{\hat{S}(x_t, u_{t:t+N-1|t})}{\lambda}) p(u_{t:t+N-1|t}),
    \end{align*}
    solve $\min_{Q} \Expectation_{Q}[\hat{S}(x_t, u_{t:t+N-1|t})] + \lambda D_{\mathrm{KL}}(Q\|P)$, where $\lambda$ denotes the inverse temperature, $\eta$ is a normalizing constant, $D_{\mathrm{KL}}$ is Kullback–Leibler divergence, and $P$ is an arbitrary base distribution with density $p(u_{t:t+N-1|t})$.
\end{proposition}

Thus, rather than directly optimizing \eqref{prob:proposed}, we seek to align the sample distribution as closely as possible with the optimal distribution by solving 
    $u_{t:t+N-1|t}^{\ast} = \argmin_{\bar{u}_{t:t+N-1|t}} D_{\mathrm{KL}}(Q^\ast \| Q_{\bar{u}_{t:t+N-1|t}, \Sigma_u})$,
which reduces to 
\begin{subequations} \label{eq:mppi_solution}
\begin{align} 
    &\hspace{-3mm} u_{t:t+N-1|t}^\ast = 
    \Expectation_{Q_{\bar{u}_{t:t+N-1|t}, \Sigma_u}}[u_{t:t+N-1|t} \omega(u_{t:t+N-1|t})], \\
    &\hspace{-3mm} \omega(u_{t:t+N-1|t}) = 
    \frac{1}{\eta}\exp(-\frac{1}{\lambda}(\hat{S}(x_t, u_{t:t+N-1|t}) \nonumber\\
    &\quad + \lambda \sum_{k=t}^{t+N-1} (u_{k|t} - \bar{u}_{k|t})^\top \Sigma_{u}^{-1} u_{k|t})).
\end{align}
\end{subequations}

\subsection{Sampling-based Implementation} \label{sec:sampling}

In order to approximate the expectations involving $a$, $b$, and $q$ in \eqref{eq:pred_belief}, \eqref{prob:proposed}, \eqref{eq:mppi_solution}, we employ importance sampling. We learn the belief distribution online using a particle filter. Initially, we generate $N_p$ samples from $b(\theta)$ according to
\begin{align} \label{eq:particle_filter_init}
    \theta^{i} \sim b_0,~i=1, \dots, N_p,
\end{align}
and initialize the corresponding weights to $\nu^i_0 = 1/N_p$.
The weights are then updated online based on the observed states using the following equations 
\begin{subequations} \label{eq:sampling_particle_filter}
\begin{align}
    \tilde{\nu}^i_{t+1} &= \nu^i_{t} h_{1}(x_{t+1} | x_t, u_t, \theta^i), \\
    \nu^i_{t+1} &= \tilde{\nu}^i_{t+1} / \sum_{i=1}^{N_p} \tilde{\nu}^i_{t+1},
\end{align}
\end{subequations}
where $h_{1}(x_{t+1} | x_t, u_t, \theta^i)$ is the conditional probability density for $x_{t+1} \sim \mathcal{N}(f(x_t, u_t, \theta^i), \Sigma_w)$ as in \eqref{eq:sys_pdf}. This function represents the particle approximation of \eqref{eq:belief_dynamics}.

\begin{remark}
    Resampling may be added to \eqref{eq:sampling_particle_filter} to enrich the sampled parameters. However, this is an implementation consideration rather than a theoretical one, and we found it unnecessary for the evaluation presented in Section~\ref{sec:application}.
\end{remark}

The predicted belief dynamics are approximated using the following predicted weights 
\begin{subequations} \label{eq:pred_weight_dynamics}
\begin{align} 
    \hat{\tilde{\nu}}^{i}_{k+1|t} &= \hat{\nu}^{i}_{k|t} \hat{h}_{k+1-t}(\bar{x}_{k+1|t} | x_t, u_{t:k|t}, \theta^i), \\
    \hat{\nu}^{i}_{k+1|t} &= \hat{\tilde{\nu}}^{i}_{k+1|t} / \sum_{i=1}^{N_p} \hat{\tilde{\nu}}^{i}_{k+1|t}.
\end{align}
\end{subequations}
Here, $\hat{h}_{k+1-t}(\bar{x}_{k+1|t} | x_t, u_{t:k|t}, \theta^i)$ is an unbiased approximation of \eqref{eq:sys_pdf} given by the conditional probability density for $\bar{x}_{k+1|t} \sim \mathcal{N}(g_{k+1-t}(x_t, u_{t:k|t}, \theta^i), \Sigma^{i}_{x_{k+1|t}})$ such that
\begin{align*}
    &\bar{x}_{k+1|t} = \sum_{i=1}^{N_p}  \nu^i_t\bar{x}^{i}_{k+1|t}, \\
    &\hspace{-1mm} \Sigma^{i}_{x_{k+1|t}} = \sum_{j=1}^{N_w} (g_{k+1-t}(x_t, u_{t:k|t}, \theta^i, w_{t:k}^j) - \bar{x}^{i}_{k+1|t})(\star)^\top / N_w \\
    &\bar{x}^{i}_{k+1|t} = \sum_{j=1}^{N_w} g_{k+1-t}(x_t, u_{t:k|t}, \theta^i, w_{t:k}^j) / N_w,
\end{align*}
where $(\star)$ represents repeated terms, $w_{t:k}^j = \{w_t^j, \dots, w_{k}^j\}$ and $w_k^j \sim a$, for $j=1, \dots, N_w$.
Thus, the objective function \eqref{prob:proposed} may be approximated by
\begin{multline}\label{eq:sampling_objective}
    \hat{\tilde{S}}(x_t, u_{t:t+N-1|t}) \\
    \approx \dfrac{1}{N_w}\sum_{i=1}^{N_p} \sum_{j=1}^{N_w} \Big[\phi(g_N(x_t, u_{t:t+N-1|t}, \theta^i, w_{t:t+N-1}^j))\hat{\nu}^i_{k+N|t} \\
    + \sum_{k=t+1}^{t+N-1} \ell(g_{k-t}(x_t, u_{t:k-1|t}, \theta^i, w_{t:k-1}^j))\hat{\nu}^i_{k|t}\Big].
\end{multline}
Finally, we approximate the optimal control \eqref{eq:mppi_solution} using $N_c$ control samples as follows 
\begin{align} \label{eq:sampling_u_ast}
    u_{t:t+N-1|t}^{\ast} &= \sum_{\ell=1}^{N_c} u_{t:t+N-1|t}^\ell \hat{\omega}(u_{t:t+N-1|t}^{\ell}),
\end{align}
where $\hat{\omega}(u_{t:t+N-1|t}^{\ell}) = \frac{1}{\eta}\exp(-\frac{1}{\lambda}(\hat{\tilde{S}}(x_t, u_{t:t+N-1|t}^\ell) + \lambda \sum_{k=t}^{t+N-1} (u_{k|t}^\ell - \bar{u}_{k|t})^\top \Sigma^{-1}_{u} u_{k|t}^\ell))$, $u_{t:t+N-1|t}^{\ell} \sim q(\cdot | \bar{u}_{t:t+N-1|t}, \Sigma_u)$ for all $\ell = 1, \dots, N_c$.

We summarize the proposed approach in Algorithm~\ref{alg:DMPPI}.

\begin{algorithm}[h]
\footnotesize
\caption{Dual Model Predictive Path-Integral Control}
\label{alg:DMPPI}
\begin{algorithmic}[1]
\Require Sample sizes and control horizon: $N_c, N_p, N_w, N$
\Require MPPI distribution parameters: $\lambda, \bar{u}_{0:N-1|0}, \Sigma_u$
\Require Parameter prior distribution: $b_0 = b(\theta)$
\State $t \gets 0$
\State $\theta^{i} \sim b_0$, $\nu_0^{i} \gets 1/N_p$, $\forall~i=1, \dots, N_p$
\Loop
\State Sample $u_{t:t+N-1|t}^{\ell} \sim q(\cdot|\bar{u}_{t:t+N-1|t}, \Sigma_u)$, $\ell=1, \dots, N_c$ 
\State $w_{t:t+N-1}^{j} \sim a$, $j=1, \dots, N_w$, $\hat{\nu}^{i}_{t|t} = \nu^{i}_t$, $i=1, \dots, N_p$
\State $\hat{\nu}^{i, \ell}_{k|t} \gets$ Evaluate \eqref{eq:pred_weight_dynamics} using $x_t, u_{t:t+N-1|t}^{\ell}, \theta^{i}, w_{t:t+N-1}^{j}$, $\forall~ \ell=1, \dots, N_c,~i=1, \dots, N_p,~j=1, \dots, N_w,~k=t+1, \dots, t+N$
\State $\hat{\tilde{S}}^{\ell} \gets$ Evaluate \eqref{eq:sampling_objective} using $x_t, u_{t:t+N-1|t}^{\ell}, \theta^{i}, \hat{\nu}^{i, \ell}_{k|t}, w_{t:t+N-1}^{j}$, $\forall~ \ell=1, \dots, N_c,~i=1, \dots, N_p,~j=1, \dots, N_w,~k=t, \dots, t+N$
\State $u_{t:t+N-1|t}^{\ast} \gets$ Evaluate \eqref{eq:sampling_u_ast} using $u_{t:t+N-1|t}^{\ell}, \hat{\tilde{S}}^{\ell}, \ell=1, \dots N_c$
\State Apply $u_t = u_{t|t}^{\ast}$ and observe $x_{t+1}$
\State $\nu_{t+1}^{i} \gets$ Evaluate \eqref{eq:sampling_particle_filter} using $\nu_{t}^{i}, x_{t+1}, x_{t}, u_{t}, \theta^{i}, i=1, \dots, N_p$
\State $t \gets t + 1$
\EndLoop
\end{algorithmic}
\end{algorithm}

\begin{theorem}
    The approximate solution to problem \eqref{prob:proposed} computed from \eqref{eq:sampling_u_ast} preserves the dual control effect \cite{hu2022active}, that is, the planned control actions affect the entropy of the predicted future belief distribution.
\end{theorem}
\begin{proof}
    The proof follows from \eqref{eq:pred_weight_dynamics}, in which the planned control sequence $u_{t:k|t}$ affects the weights $\hat{\tilde{\nu}}^{i}_{k+1|t}$ of the categorical distribution over $\{\theta^i\}_{i=1}^{N_p}$.
\end{proof}

\begin{remark}
    The sampling complexity of computing the optimal control from \eqref{eq:sampling_u_ast} is $\mathcal{O}(N_cN_p N_w N)$. Therefore, the sample sizes and control horizon must be carefully chosen to balance the accuracy of the sampling approximations of the respective distributions with practical computational and memory constraints.
\end{remark}

\section{Autonomous Highway On-ramp Merging Application} \label{sec:application}

\begin{figure}[h]
    \centering
    \includegraphics[width=\columnwidth]{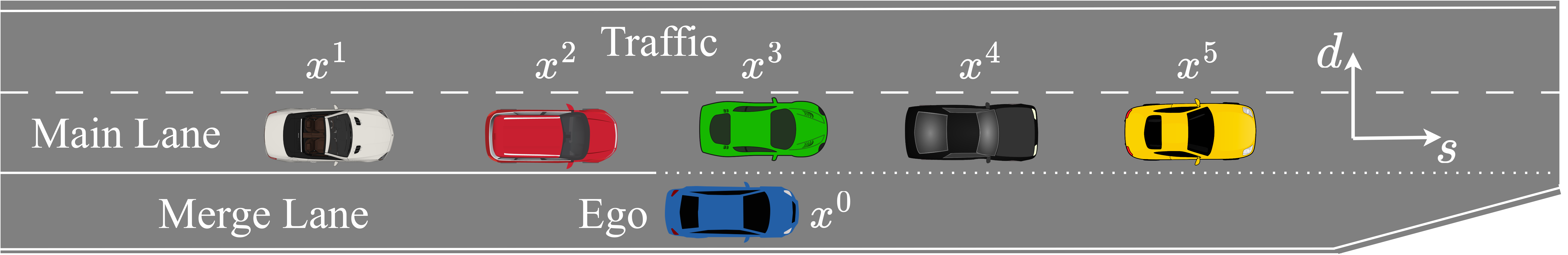}
    \caption{Highway on-ramp merge scenario. Ego car is the autonomous merging car (in blue color) while all the main lane traffic cars are human driven vehicles.}
    \label{fig:merge_scenario}
\end{figure}

We evaluate the proposed approach \eqref{prob:proposed} using the MPPI solution method \eqref{eq:mppi_solution} in a challenging highway on-ramp merge scenario featuring dense traffic conditions shown in Fig.~\ref{fig:merge_scenario}. 
We use the DMPPI framework for interaction-aware \emph{trajectory generation}, while low-level control is handled by the Vehicle Control System (VCS) of the Autonomous merging vehicle, also referred to as the ego vehicle.  Therefore, the purpose of the DMPPI framework is primarily to design trajectories that will enable learning the traffic vehicles' behavior parameters and lead to a successful merge, while generating actuator commands to follow such a trajectory is left to the VCS.

We consider a highway on-ramp merging scenario with a single ego vehicle in the merge lane and $n_v$ vehicles in the main road, where $1$ and $n_v$ denote the indices of the rear and lead vehicles, respectively. 
Let $0$ be the index of the ego vehicle in the merge lane.
At time step $k\in\mathbb{N}$, let $x_k^i=[v_{s,k}^i, v_{d,k}^i, s_k^i, d_k^i]^\top \in\mathbb{R}^4$ be the vector corresponding to the state of vehicle $i\in\{0,1,\dots,n_v\}$, where $v_{s,k}^i$, $v_{d,k}^i$, $s_k^i$, and $d_k^i$ denote longitudinal speed, lateral speed, longitudinal position, and lateral position, respectively in the Frenet coordinate frame. 
Let the stacked state vector of all vehicles at time step $k$ be given by $x_k=[x_k^0, x_k^1, \cdots, x_k^{n_v}]^\top$. 
The control input of the ego vehicle is denoted by $u_k$. 
We are interested in high-level, long-horizon motion planning, and assume that more complex low-level dynamics and control be handled by the autonomous vehicle's VCS. 
We therefore consider a double integrator dynamics, and let the control input of the ego vehicle be equal to its acceleration in the local Frenet coordinate frame, i.e., $u_k=[\hat{\dot{v}}_{s,k}^0, \hat{\dot{v}}_{d,k}^0]^\top\in\mathbb{R}^2$. 

To predict how the vehicles on the main road behave, we employ the Merge Reactive Intelligent Driver Model (MR-IDM) \cite{holley2023mr}. This model extends the widely-used IDM \cite{treiber2000congested} by also considering reaction to the merging vehicles in addition to the following vehicles in computing the longitudinal acceleration of the driver. We assume that main road vehicles do not make any lane changes, and thus, we only consider their longitudinal motion. 

The combined ego-traffic dynamics are given by \eqref{eq:sys} with 
\begin{align} \label{vehicles_model}
    &\hspace{-4mm}f(x_t, u_t, \theta) \! = \!
    \mathcal{A}
    x_t 
    \!+ \!
    \mathcal{B} \!
    \begin{bmatrix}
        \sigma(x_t, u_t) \\
        \zeta^{1}(x^1_t, x^0_t, x^{2}_t, \theta^1) \\
        \vdots \\
        \zeta^{n_v-1}(x^{n_v-1}_t, x^0_t, x^{n_v}_t, \theta^{n_v-1}) \\
        \zeta^{n_v}(x^{n_v}_t, x^0_t, \theta^{n_v})
    \end{bmatrix},
\end{align}
where $\mathcal{A} = \mathrm{blkdiag}(A, \dots, A)$, $\mathcal{B} = \mathrm{blkdiag}(B, \dots, B)$,
\begin{subequations} 
\begin{align}
    &\zeta^{m} = \begin{bmatrix}
        \dot{v}_s^{m} \\
        \dot{v}_d^{m}
    \end{bmatrix} = \begin{bmatrix}
        \rho(x^m_t, x^0_t, x^{m+1}_t, \theta^m)\\
        0
    \end{bmatrix}, \\
    &\zeta^{n_v} = \begin{bmatrix}
        \dot{v}_s^{n_v} \\
        \dot{v}_d^{n_v}
    \end{bmatrix} = \begin{bmatrix}
        \tilde{\rho}(x^{n_v}_t, x^0_t, \theta^{n_v})\\
        0
    \end{bmatrix},\\
    &A = \begin{bmatrix}
        1 & 0 & 0 & 0 \\
        0 & 1 & 0 & 0 \\
        \Delta t & 0 & 1 & 0 \\
        0 & \Delta t & 0 & 1
    \end{bmatrix},~B = \begin{bmatrix}
        \Delta t & 0 \\
        0 & \Delta t \\
        0 & 0 \\
        0 & 0
    \end{bmatrix},
\end{align}
\end{subequations}
for $m\in\{1,2, \dots, n_v-1\}$. Let $\sigma(x_t, u_t) = [\dot{v}_{s,k}^0, \dot{v}_{d,k}^0]^{\top}$ in \eqref{vehicles_model} be a clamping function which imposes kinematic constraints such as requiring $v^{0}_s \geq 0$ and limiting acceleration magnitudes ($\dot{v}_{s_{\text{min}}}^0 \leq \dot{v}_{s,k}^0 \leq \dot{v}_{s_{\text{max}}}^0$, $\dot{v}_{d_{\text{min}}}^0 \leq \dot{v}_{d,{k}}^0 \leq \dot{v}_{d_{\text{max}}}^0$). 
Let $\zeta^{i}_t$ be the acceleration of traffic vehicle $i$ for $i\in\{1,\dots,n_v\}$, while $\rho(\cdot)$ and $\tilde{\rho}(\cdot)$ are the MR-IDM with and without a lead vehicle, respectively.\footnote{The equations for $\rho(\cdot)$ and $\tilde{\rho}(\cdot)$ are given in \cite{holley2023mr}.} 
Note that $\zeta^{n_v}$ does not depend on a third vehicle, as $x^{n_v}$ corresponds to the front-most traffic vehicle. 

We set the cost function \eqref{eq:cost_function} at time step $t$ for the control horizon of length $N\in\mathbb{Z}^+$ according to 
\begin{subequations}
\begin{align}
    \ell_k(x_k) =& (x_k - x^g)^\top Q (x_k - x^g) + \ell^\text{penalty}(x_k), \\
    \phi(x_{t+N}) =& (x_{t+N} - x^g)^\top Q_f (\star) + \ell^\text{penalty}(x_{t+N}), \\
    \ell^\text{penalty}(x_k) &= q_I(\mathbf{1}^{\text{coll}}(x_k) + \mathbf{1}^\text{road}(x_k) + \mathbf{1}^\text{inval}(x_k)),
\end{align}
\end{subequations}
for $k=\{t+1, t+2,\dots,t+N-1 \}$,
where $x^g = [v^g, 0, 0, 0]^\top$ is the goal state, $Q = \text{diag}(q_{v_s}, q_{v_d}, q_{s}, q_{d}, 0, \dots, 0) \in \Re^{4(n_v+1) \times 4(n_v+1)}$ is the state cost matrix, $Q_f = \text{diag}(0, 0, 0, q^f_{d}, 0, \dots, 0) \in \Re^{4(n_v+1) \times 4(n_v+1)}$ is the terminal cost matrix, $q_{I}$ is the violation penalty coefficient, and $\mathbf{1}^{\text{coll}}(x_k)$, $\mathbf{1}^\text{road}(x_k)$, $\mathbf{1}^\text{inval}(x_k)$ are the indicator functions for a collision with another vehicle, violating the road boundaries, or an improper merge (not between two vehicles), respectively. 
The selected parameters are shown in Table~\ref{tab:parameters}. The cost function is designed to prioritize avoiding violations of the indicator functions, merging into the main lane of traffic, and maintaining the desired velocity.

\begin{table}[h]
\renewcommand{\arraystretch}{1.15}
\centering
\caption{DMPPI Parameters}
\label{tab:parameters}
\begin{tabular}{|l|l|l|}
\hline
Parameter & Value & Description \\ \hline
$N_c$ &$ 3,\!000$ & \# of control samples \\ \hline
$N^{pf}_p$ & $10,\!000$ & \# of parameter samples for particle filter\\ \hline
$N^{c}_p$ & $20$ & \# of parameter samples for control \\ \hline
$N_w$ & $5$ & \# of disturbance samples \\ \hline
$N$ & $50$ & \# of control horizon time-steps \\ \hline
$\lambda$ & $10,\!000$ & Sharpness of MPPI optimal distribution \\ \hline
$\Sigma_u$ & diag($10$, $1.5$)  & Control sample covariance  (m/s$^2$)$^2$\\ \hline
$v^g $ & $10$& Target velocity m/s \\ \hline
$q_{v_s}$ & $10$ & Longitudinal velocity cost \\ \hline
$q_{v_d}$ & $0.1$ & Lateral velocity cost \\ \hline
$q_{s}$ & $0$ & Longitudinal position cost \\ \hline
$q_{d}$ & $10$ & Lateral position cost \\ \hline
$q_{d}^f$ & $10,\!000$ & Terminal lateral position cost \\ \hline
$q_{I}$ & $1,\!000,\!000$ & Violation penalty \\ \hline
\end{tabular}
\end{table}

\begin{remark}
    An important implementation consideration is that, whereas the particle filter \eqref{eq:particle_filter_init}-\eqref{eq:sampling_particle_filter} and the predicted belief dynamics used for dual control \eqref{eq:pred_weight_dynamics}-\eqref{eq:sampling_objective} ideally utilize the same number of parameter samples $N_p$, in practice we use $N_p^{pf}$ samples for the particle filter and $N_p^c$ samples for control. 
    We employ resampling to reduce the number of samples from the particle filter (which is relatively cheap) to DMPPI (which is more computationally expensive).
\end{remark}

We first evaluated the proposed approach in a low-fidelity highway on-ramp simulation environment that was developed in-house. 
We setup the evaluation to only include challenging merge scenarios requiring negotiation wherein the ego vehicle has less than $20$ seconds and less than $300$ meters from the soft-nose to the merge-ramp endpoint to successfully merge between five traffic vehicles, as shown in Fig.~\ref{fig:merge_scenario}. 
The traffic vehicles are initialized with $v_{s_0}^m = 10$~m/s, {$v_{d_0}^m = 0$~m/s}, $d_0^m = 0$~m, and $s_0^{m+1} = s_0^{m} + 8$, where $s_0^{1} = 0$, for $m = 1, \dots, n_v$. 
That is, all traffic vehicles begin travelling at $10$~m/s, $8$~m apart from one another.
The ego vehicle is initialized with $v_{s,0}^0 = 10$~{m/s}, $v_{d,0}^0 = 0$~m/s, $d_0^0 = -3.5$~m, and $s_0^{0} \sim \mathcal{U}(s_0^1, s_0^{n_v})$, where $\mathcal{U}(s_0^1, s_0^{n_v})$ is a uniform distribution over the interval $[s_0^1, s_0^{n_v}]$.
To robustly evaluate the proposed approach in comparison to the baselines, we randomly assign only one of the traffic vehicles to have a relatively friendly set of parameters of the MR-IDM, $\bar{\theta}^{\tilde{m}} \sim \Theta^{\text{friendly}}$, where $\tilde{m} \sim \{1, 2, \dots, n_v\}$, thereby resulting in a scenario in which this specific vehicle will yield to the ego vehicle in case it attempts to merge in front of it. 
The rest of the vehicles have aggressive parameters, $\bar{\theta}^{\bar{m}} \sim \Theta^{\text{aggressive}}$, where $\bar{m} \neq \tilde{m}$, and will not yield. 
To further increase the difficulty of the scenario, merging behind the last car or ahead of the first car was treated as a failure case.

The proposed approach was compared against two state-of-the-art sampling-based baselines that do not feature active learning, but otherwise utilize the same particle filter and MPPI control solution approaches: CE-MPPI and EMPPI, where the objective $\hat{S}$ in \eqref{eq:sampling_u_ast} is replaced by the objectives of \eqref{prob:ce_mppi_ocp} and \eqref{prob:emppi_ocp}, respectively. 
All approaches use the same model, cost function, and parameters, where applicable. 
An important consideration is the design of the prior belief distribution over the parameters which we set to $b_0 = b(\theta) = 0.8\,\Theta^{\text{friendly}} + 0.2\,\Theta^{\text{aggressive}}$.
The merge success rate result for this low fidelity simulation test is shown in Table~\ref{tab:results}.

\begin{table}[h]
\centering
\caption{Monte Carlo Merge Simulation Results}
\label{tab:results}
\begin{tabular}{|l||l|l|l|}
\hline
Method       & CE-MPPI & EMPPI & \textbf{DMPPI} \\ \hline
Low-Fidelity Success Rate & $0.44$    & $0.47$  & $0.70$            \\ \hline
High-Fidelity Success Rate & $0.37$    & $0.31$  & $0.67 $          \\ \hline
Avg. Min Long. Distance (m) & $1.74$ & $1.45$ & $1.90$ \\ \hline
Avg. Min Lat. Distance (m) &$ 1.07$ & $1.06$ & $1.14$ \\ \hline
Avg. Max Acceleration (m/s$^2$) & $1.89$ & $1.21$ & $2.55$ \\ \hline
\end{tabular}
\end{table}

\begin{figure}[h]
    \centering
    \begin{subfigure}{0.48\columnwidth}
        \includegraphics[width=\textwidth]{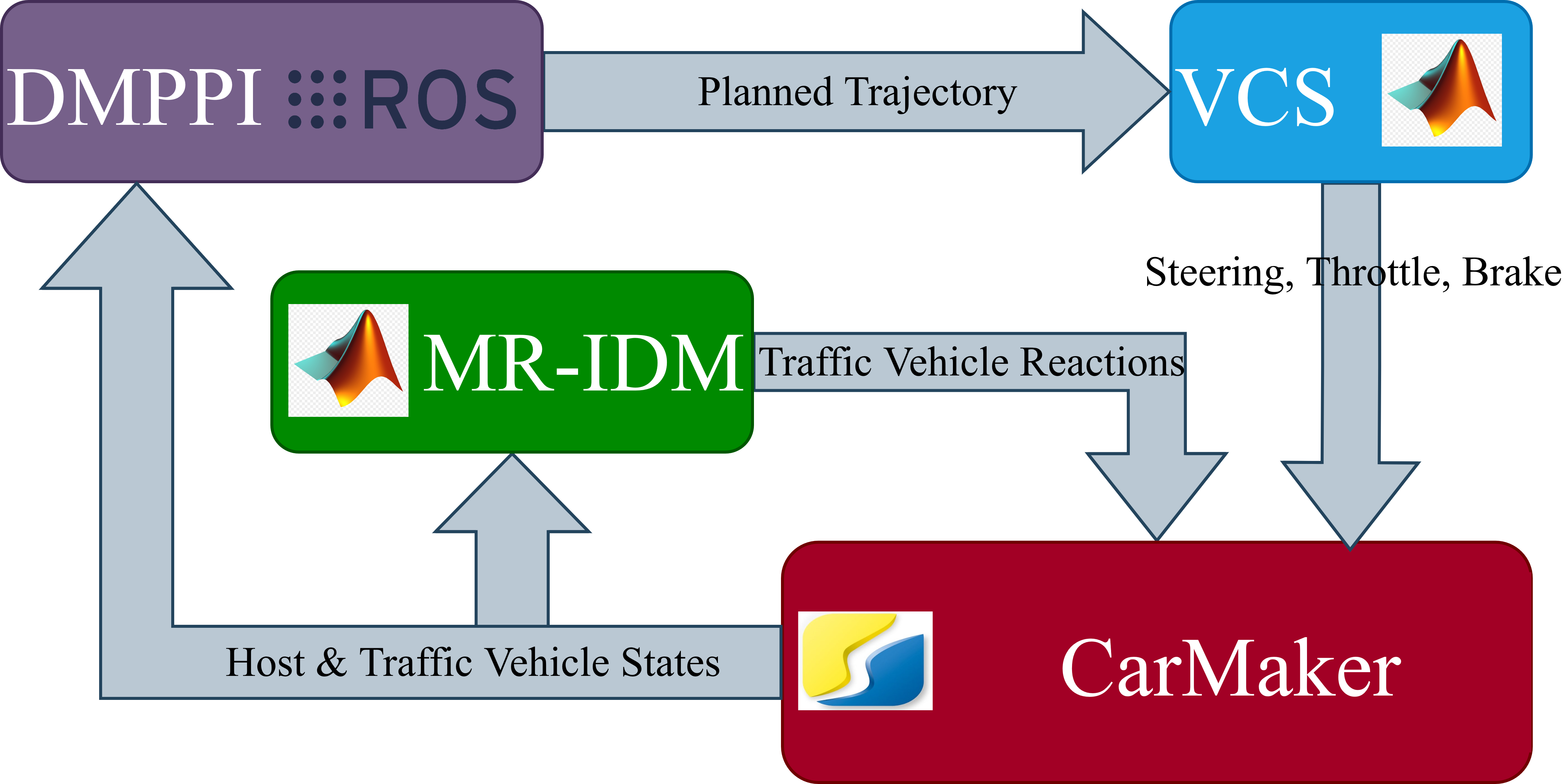}
    \end{subfigure}
    \begin{subfigure}{0.48\columnwidth}
        \includegraphics[width=\textwidth]{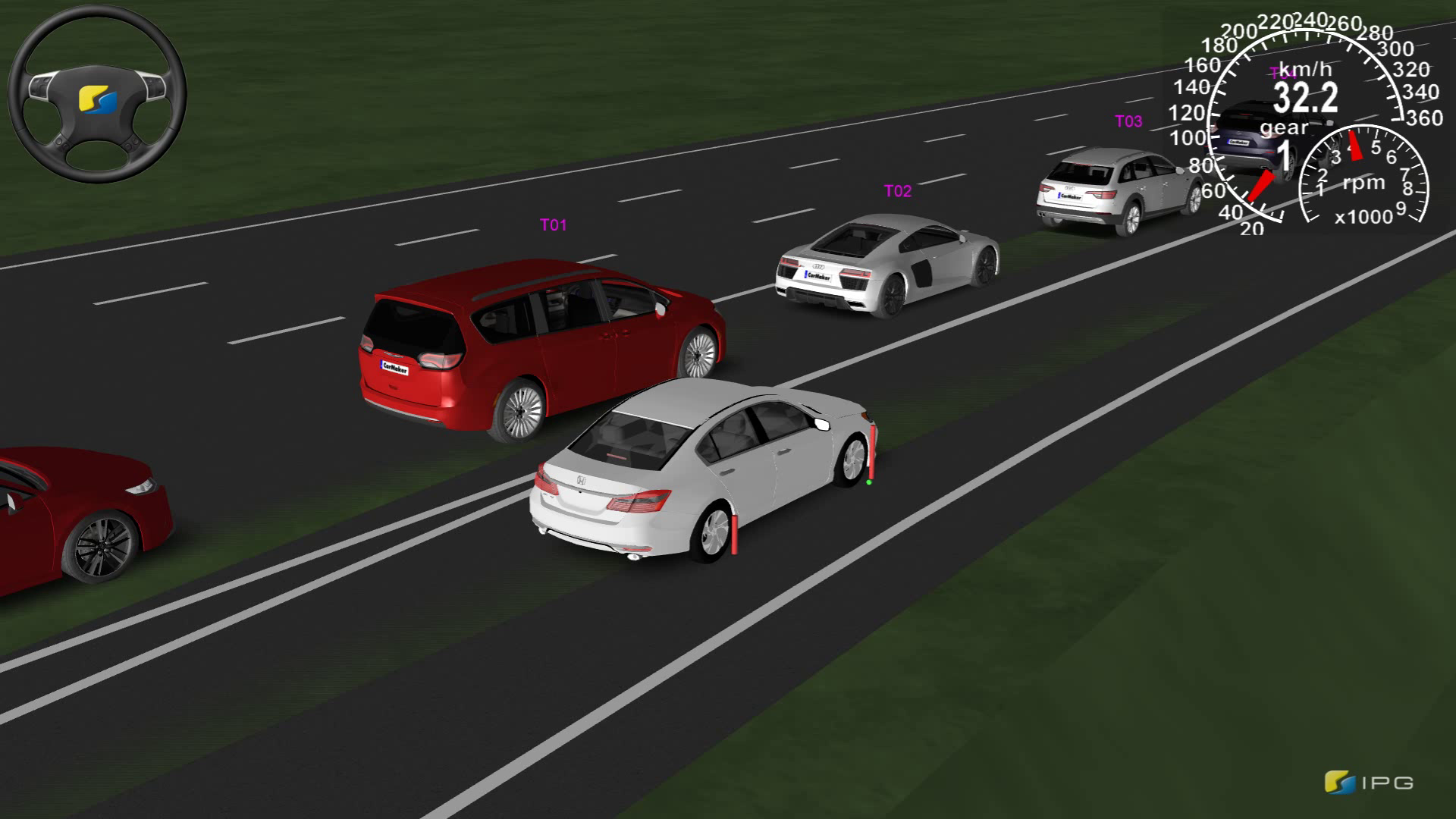}
    \end{subfigure}
    \caption{IPG-CarMaker high-fidelity simulation environment.}
    \label{fig:ipg_carmaker}
\end{figure}

We next proceeded to evaluate the performance of the proposed approach in a real-time high-fidelity simulation environment using IPG-CarMaker \cite{carmaker2021reference} and ROS (Robot Operating System), as shown in Fig.~\ref{fig:ipg_carmaker}. 
We setup the traffic similar to the low-fidelity simulation setup to replicate the difficult merge scenarios. 
However, the highway used in this evaluation had a soft-nose to ramp-end distance of only $140$ meters. 
To offset this added difficulty, we allowed up to two traffic vehicles to have parameters sampled from $\Theta^{\text{friendly}}$.
The DMPPI algorithm ran in real-time at approximately $10$ Hz on a single NVIDIA RTX $3080$ Ti GPU using Jax \cite{jax2018github} for just-in-time compilation and hardware acceleration. 

\begin{figure}[h]
    \centering
    \begin{subfigure}{0.98\columnwidth}
        \includegraphics[width=0.99\textwidth]{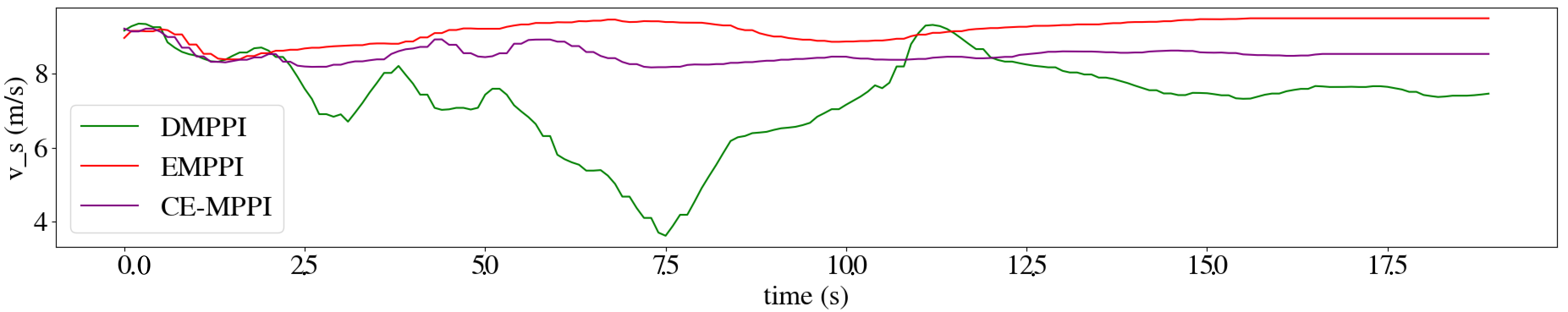}
    \end{subfigure}
    \begin{subfigure}{0.31\columnwidth}
        \includegraphics[width=\textwidth, trim={15 50 50 150},clip]{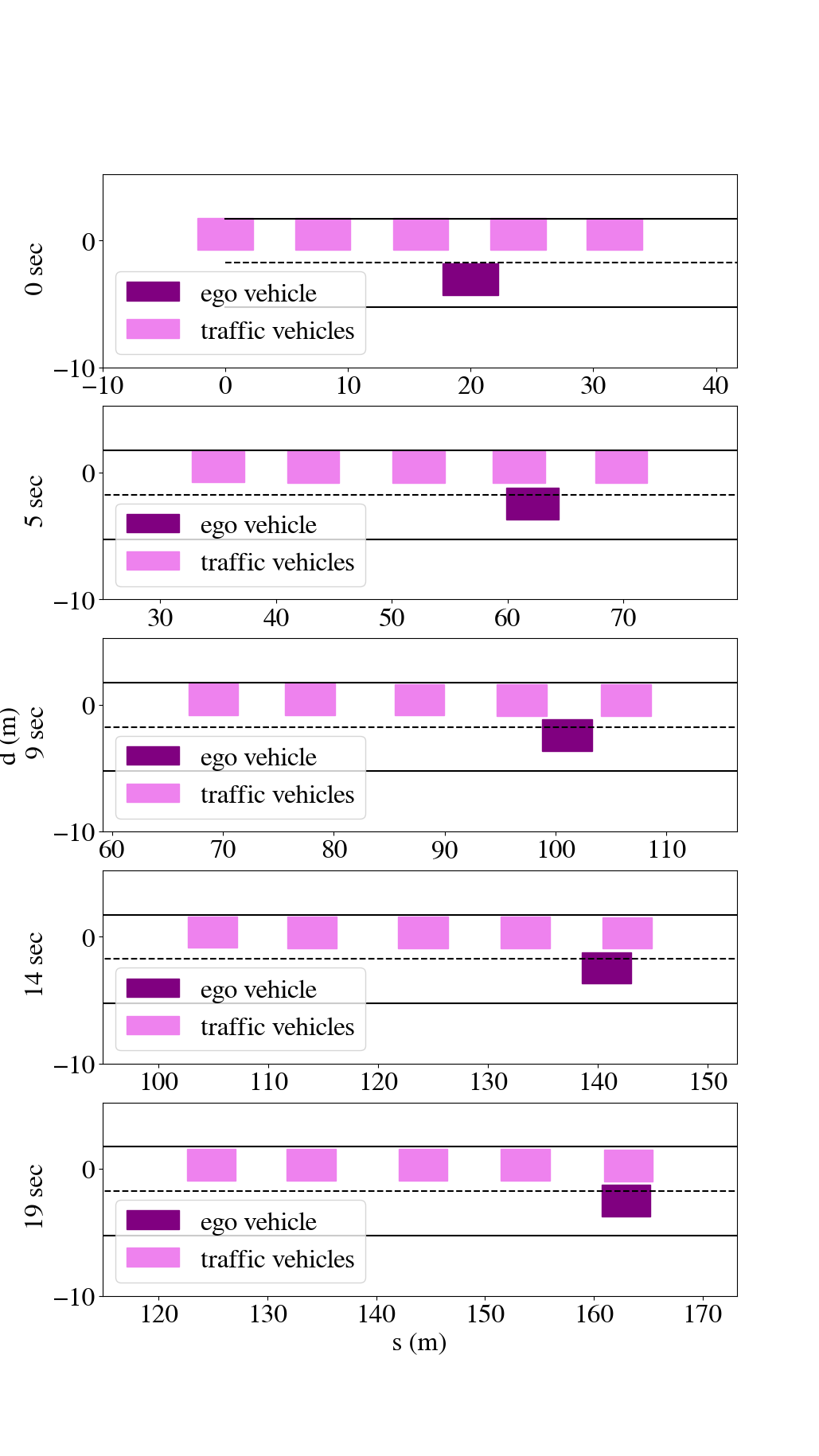}
        \caption{CE-MPPI}
    \end{subfigure}
    \begin{subfigure}{0.31\columnwidth}
        \includegraphics[width=\textwidth, trim={15 50 50 150},clip]{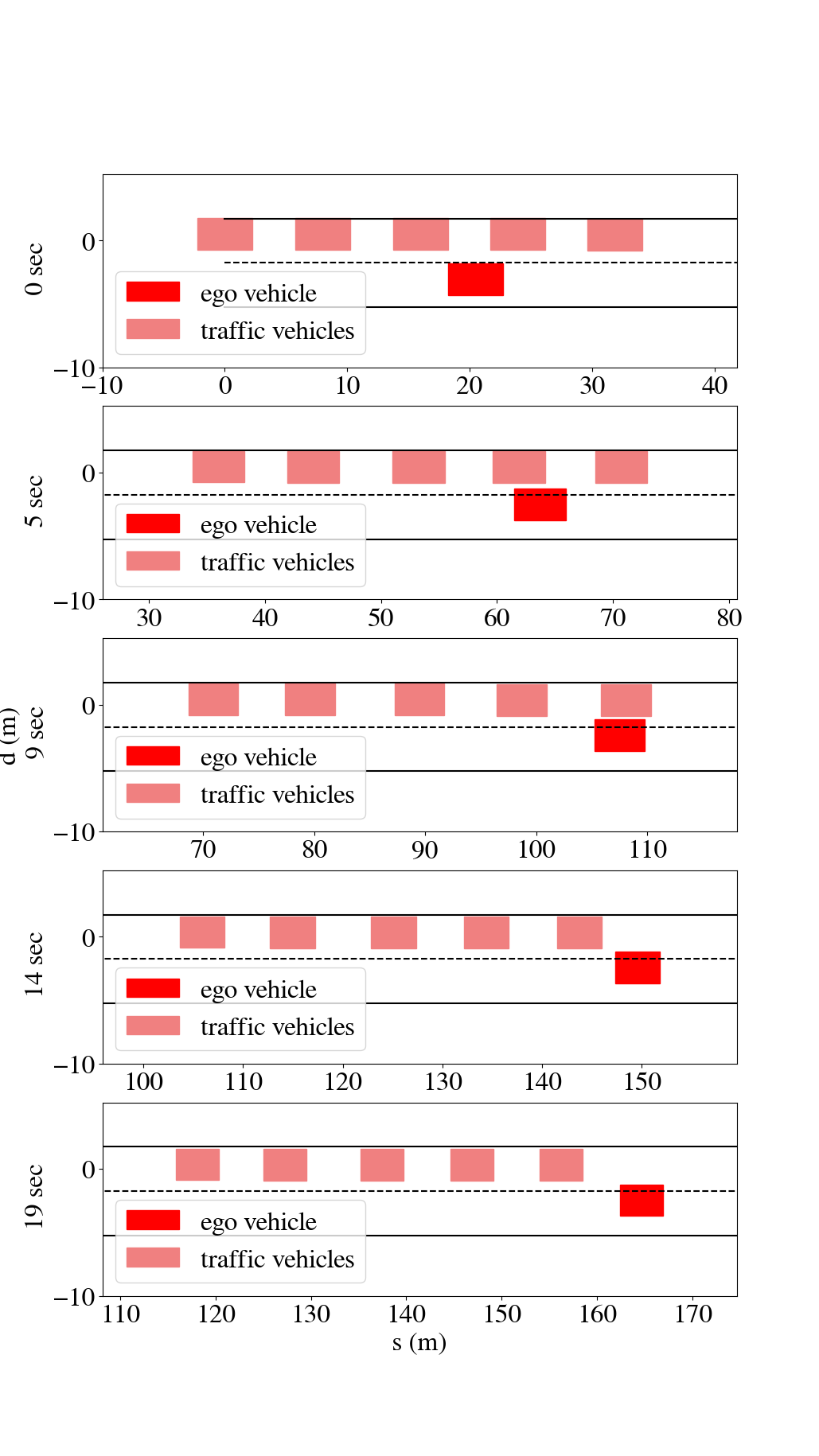}
        \caption{EMPPI}
    \end{subfigure}
    \begin{subfigure}{0.31\columnwidth}
        \includegraphics[width=\textwidth, trim={15 50 50 150},clip]{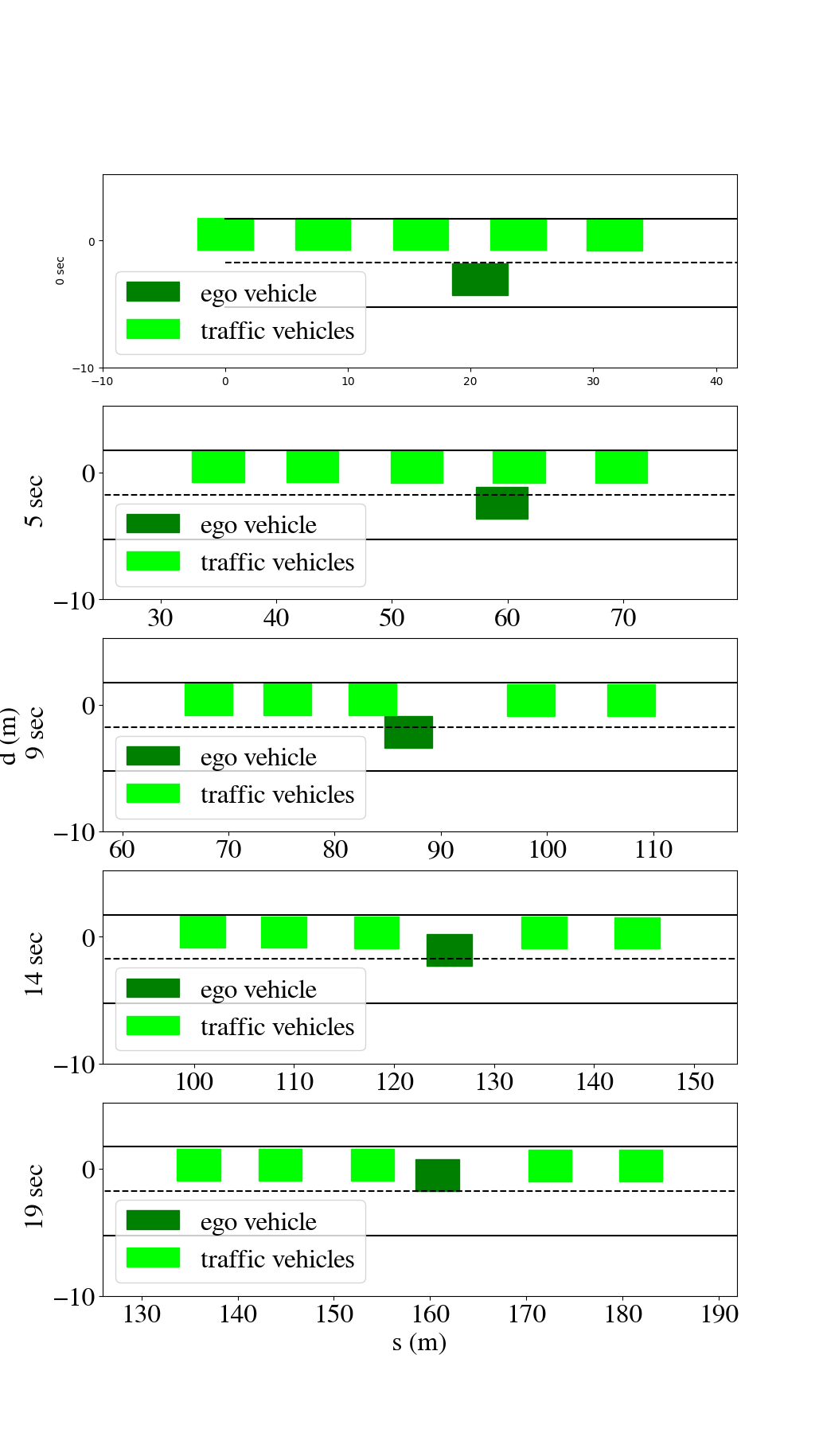}
        \caption{DMPPI}
    \end{subfigure}
    \caption{Comparison of snapshots of trajectories and velocity profiles from a single Monte Carlo merge scenario trial.}
    \label{fig:pyplot_snapshots_4_1}
\end{figure}

The DMPPI control algorithm generates a planned trajectory using \eqref{vehicles_model}. Subsequently, this trajectory is broadcast via ROS to the VCS, which operates in the MATLAB-Simulink environment. 
The VCS computes the low-level control actions to attempt follow the planned trajectory and sends them over ROS to CarMaker which then simulates the effects of these control actions on the ego vehicle. 
Additionally, the behavior of the traffic vehicles characterized by MR-IDM is processed within the MATLAB environment and the responses are broadcast to CarMaker over ROS, as shown in Fig.~\ref{fig:ipg_carmaker}. 
The proposed approach was again compared against CE-MPPI and EMPPI. 
The rates of a successful merge (out of $100$ trials) for each of the three approaches are shown in Table~\ref{tab:results}, and a comparison of a single trial may be seen in Fig.~\ref{fig:pyplot_snapshots_4_1}.

Although DMPPI has a higher success rate over the baselines, it does not compromise safety as all three methods attained a $0$\% collision rate and maintained similar distances from the traffic vehicles as seen in Table~\ref{tab:results}, where the 4\textsuperscript{th} and 5\textsuperscript{th} rows refer to the minimum longitudinal and lateral distances between the ego vehicle and any traffic vehicle averaged over all Monte Carlo trials.
Rather, the superior performance is attained by intelligently expending more control effort in order to probe the system as seen in the last row of Table~\ref{tab:results}, which refers to the average over the Monte Carlo trials of the maximum absolute acceleration of the ego vehicle for each trial. This resulting higher acceleration for DMPPI is still lower than the common acceleration limits for AD/ADAS systems \cite{de2023standards}.

The relative performance between the three approaches is comparable to that for the low-fidelity simulation. 
However, the overall performance for all approaches is slightly reduced in the high-fidelity simulation, which is due, in part, to the model mismatch between the environment and the model used for control design and parameter fitting as well as due to the additional kinematic and dynamic limitations imposed by the high-fidelity vehicle model.

The explanation for the superior merge success rate of DMPPI may be seen through Fig.~\ref{fig:pyplot_snapshots_4_1}, which shows snapshots at different time instants for a single trial.
Both CE-MPPI and EMPPI algorithms tend to maintain a relatively constant velocity since no merging gap is available and these observations fail to provide any indication of a forthcoming gap opening up for merging. 
DMPPI, on the other hand, actively probes the vehicles, by adjusting its velocity, in order to better identify the driver's behavior and find a friendly driver in front of whom to merge. 
It may be seen that, initially, all three approaches follow the same trajectory. However, once it is seen that a driver will not yield, in this case, DMPPI sharply slows down to probe another driver, while the other two approaches maintain their velocity and relative positions to the traffic vehicles waiting for a feasible gap to open.
By decelerating, DMPPI shifts its position to better identify the parameters of another driver, hoping that they will be friendly and willing to yield.
By doing so, a yielding vehicle is found and the merge may be completed.

\section{Conclusion}
In this paper, we introduced the novel Dual Model Predictive Path-Integral (DMPPI) control algorithm, a sampling-based implicit dual optimal control framework. 
We employ this framework in a highway on-ramp merge scenario in which our algorithm is used to actively learn the unknown behavior of other drivers in order to successfully merge into congested freeway traffic.
We evaluated the performance of our framework in a high-fidelity simulation environment with the algorithm running in real-time at $10$ Hz.
The proposed approach demonstrated superior performance over two state-of-the-art variations of MPPI that rely on passive learning.
Future work may include more rigorous integration between MPPI and the VCS, evaluation on more complex on-ramps and traffic scenarios generated from real-life conditions, and deployment on a physical autonomous vehicle.

\bibliographystyle{IEEEtran}
\bibliography{references}

\end{document}